\pdfoutput=1

\documentclass[11pt]{article}

\usepackage{acl}

\usepackage{times}
\usepackage{latexsym}
\usepackage[T1]{fontenc}
\usepackage[utf8]{inputenc}
\usepackage{microtype}
\usepackage{xspace}
\usepackage{enumitem}
\usepackage{booktabs}
\usepackage{amsmath}
\usepackage{amssymb}
\usepackage{pifont}
\usepackage{multirow}
\usepackage{tablefootnote}
\usepackage{graphicx}

\newcommand{\cmark}{\ding{51}}%
\newcommand{\xmark}{\ding{55}}
\newcommand{\mrtydi}{Mr.~TyDi\xspace}

\newcommand{\ignore}[1]{}

\newcommand\Ar{\texttt{ar}\xspace}
\newcommand\Bn{\texttt{bn}\xspace}
\newcommand\En{\texttt{en}\xspace}
\newcommand\Fi{\texttt{fi}\xspace}
\newcommand\Id{\texttt{id}\xspace}
\newcommand\Ja{\texttt{ja}\xspace}
\newcommand\Ko{\texttt{ko}\xspace}
\newcommand\Ru{\texttt{ru}\xspace}
\newcommand\Sw{\texttt{sw}\xspace}
\newcommand\Te{\texttt{te}\xspace}
\newcommand\Th{\texttt{th}\xspace}

\newcommand{\insertlanguageinfotable}{
    \begin{table}[t]
     \centering
    \resizebox{0.5\textwidth}{!}{
      \begin{tabular}{l|l|l|l}
        \hline
        \textbf{Language} & \textbf{ISO} & \textbf{Language Family} & \textbf{Lucene Analyzer}\\
        \hline
        Arabic & \Ar & Afro-Asiatic & ArabicAnalyzer \\
        Bengali & \Bn & Indo-European & BengaliAnalyzer \\
        English & \En & Indo-European & EnglishAnalyzer \\
        Finnish & \Fi & Uralic & FinnishAnalyzer\\
        Indonesian & \Id & Austronesian & IndonesianAnalyzer \\
        Japanese & \Ja & Japonic & JapaneseAnalyzer \\
        Korean & \Ko & Koreanic & KoreanAnalyzer\\
        Russian & \Ru & Indo-European & RussianAnalyzer\\
        Swahili & \Sw & Niger-Congo & --\\
        Telugu & \Te & Dravidian &  TeluguAnalyzer \\
        Thai & \Th & Kra-Dai & ThaiAnalyzer \\
        \bottomrule
      \end{tabular}
      }
      \caption{
      \textbf{Language and analyzer information}: 
      The ISO-639 code, language family, and the corresponding Lucene (v9.3.0) analyzer for each language used in our experiments.
      Note that all languages except for  Swahili (\Sw) have a custom language-specific analyzer. 
      }
      \label{tab:analyzer}
    \end{table}
}

\newcommand{\insertmrtydiresults}{
\begin{table*}[!t]
     \begin{center}
     \small
      \begin{tabular}{lccccccccccc}
        \toprule[1.5pt]
        \textbf{Tokenizer} & \textbf{\Ar} & \textbf{\Bn} & \textbf{\En} & \textbf{\Fi} & \textbf{\Id}  & \textbf{\Ja}  & \textbf{\Ko}  & \textbf{\Ru} & \textbf{\Sw} & \textbf{\Te}& \textbf{\Th}  \\ 
        \midrule
        Latin script? & \xmark & \xmark & \cmark & \cmark & \cmark & \xmark & \xmark & \xmark & \cmark & \xmark & \xmark  \\
        \midrule
        & \multicolumn{10}{c}{\textbf{MRR@100}} & \\
        (1) w/ whitespace & 0.183 & 0.354 & 0.078 & 0.159 & 0.207 & 0.007 & 0.221 & 0.149 & 0.389 & 0.343 & 0.191 \\ 
        (2) w/ Lucene Analyzer & 0.368 & 0.418 & 0.140 & 0.284 & 0.376 & 0.213 & 0.285 & 0.316 & -- & 0.528 & 0.401 \\ 
        (3) w/ mBERT & 0.251 & 0.245 & 0.134 & 0.264 & 0.297  & 0.179 & 0.169 & 0.327 & 0.420 & 0.067 & 0.006 \\ 
        (4) Fusion (2 \& 3$^*$) & 0.368 & 0.409 & 0.149 & 0.305 & 0.366  & 0.232 & 0.284 & 0.349 & 0.422 & 0.272 & 0.246 \\ 
        \midrule
        & \multicolumn{10}{c}{\textbf{Recall@100}} & \\
        (1) w/ whitespace & 0.489 & 0.730 & 0.327 & 0.480 & 0.546 & 0.015 & 0.396 & 0.302 & 0.764 & 0.758 & 0.513 \\
        (2) w/ Lucene Analyzer & 0.793 & 0.869 & 0.537 & 0.720 & 0.843  & 0.643 & 0.619 & 0.654 & -- & 0.897 & 0.853 \\
        (3) w/ mBERT& 0.634 & 0.689 & 0.477 & 0.739 & 0.739 & 0.571 & 0.367 & 0.672 & 0.812 & 0.239 & 0.036 \\
        (4) Fusion (2 \& 3$^*$) & 0.793 & 0.892 & 0.538 & 0.766 & 0.838 & 0.664 & 0.589 & 0.745 & 0.842 & 0.714 & 0.817 \\
        \bottomrule[1.5pt]
      \end{tabular}
      \vspace{0.2cm}
      \caption{Results of BM25 on the test set of \mrtydi, using different tokenization mechanisms:\
      (1) whitespace, (2) a language-specific Lucene analyzer, (3) the mBERT tokenizer, and (4) a fusion of (2) and (3). For Swahili (\Sw), Lucene does not provide a custom language-specific analyzer.
      Latin-script languages are marked with~\cmark, otherwise with~\xmark. \\
      {\footnotesize $^*$For \Sw, fusion is based on (1) and (3), as (2) is not available.} 
      }
      \label{tab:evalres}
      \end{center}
\end{table*}}

\newcommand{\insertmrtydibarplot}{
\begin{figure*}[!t]
    \centering
    \includegraphics[width=\textwidth]{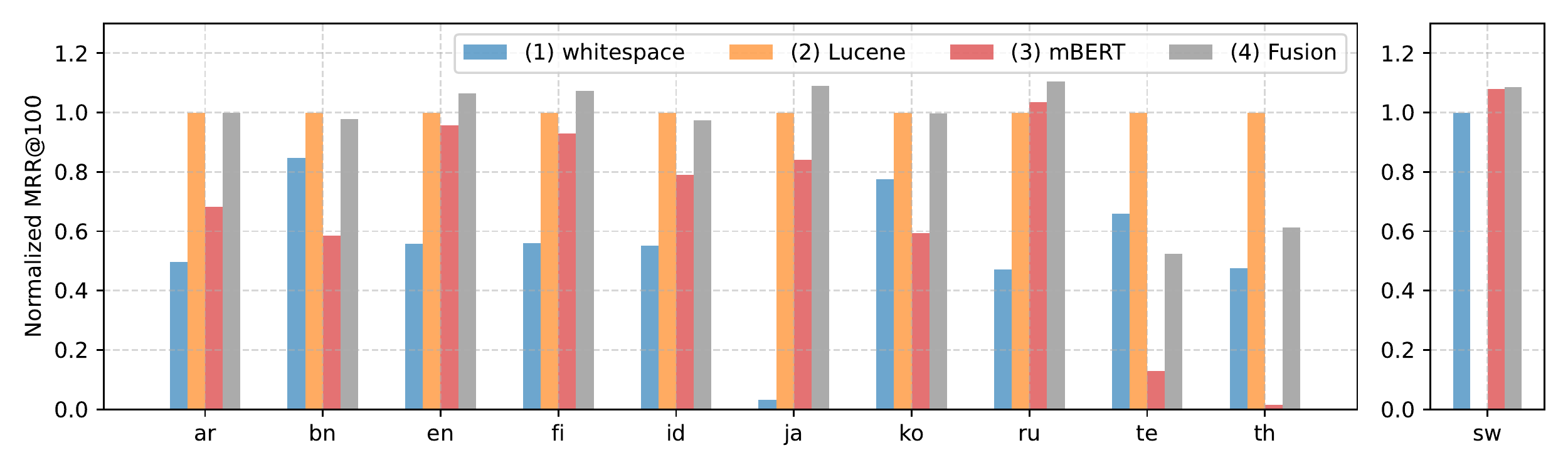}
    \caption{
     Bar chart of normalized MRR@100 from Table~\ref{tab:evalres}.
     Scores of all languages except for \Sw (left) are normalized based on the MRR@100 of the Lucene analyzer,
     with the normalized scores of the Lucene analyzer set to $1.0$ (\textbf{\textcolor{orange}{orange}} bars).
     Scores of \Sw (right) are normalized based on the MRR@100 of the whitespace tokenizer, 
     with the normalized score of whitespace tokenization set to $1.0$ ({\bf \textcolor{blue}{blue}} bars).
    }
    \label{fig:mrtydiMRR}
\end{figure*}
}

\usepackage[hang,flushmargin]{footmisc}

\title{Better Than Whitespace: Information Retrieval for Languages\\ without Custom Tokenizers}

\author{
Odunayo Ogundepo, 
Xinyu Zhang, \and
Jimmy Lin \\[1ex]
David R. Cheriton School of Computer Science\\
University of Waterloo 
}

\begin{document}
\maketitle
\begin{abstract}

Tokenization is a crucial step in information retrieval, especially for lexical matching algorithms, where the quality of indexable tokens directly impacts the effectiveness of a retrieval system.
Since different languages have unique properties, the design of the tokenization algorithm is usually language-specific and requires at least some lingustic knowledge.
However, only a handful of the 7000+ languages on the planet benefit from specialized, custom-built tokenization algorithms, while the other languages are stuck with a ``default'' whitespace tokenizer, which cannot capture the intricacies of different languages.
To address this challenge, we propose a different approach to tokenization for lexical matching retrieval algorithms (e.g., BM25):\ using the WordPiece tokenizer,
which can be built automatically from unsupervised data.
We test the approach on 11 typologically diverse languages in the \mrtydi collection:\ results show that the mBERT tokenizer provides strong relevance signals for retrieval ``out of the box'', outperforming whitespace tokenization on most languages.
In many cases, our approach also improves retrieval effectiveness when combined with existing custom-built tokenizers. 
\end{abstract}

\section{Introduction}

A fundamental assumption in information retrieval (IR) is the existence of some mechanism that converts documents into sequences of tokens, typically referred to as \textit{tokenization}.
These tokens comprise the index terms that are used to compute query--document scores when matching search queries to relevant documents in lexical matching techniques such as BM25 \citep{bm25_bibtex}.

Some of the operations involved in tokenization for the purposes of IR include case folding, normalization, stemming, lemmatization, stopwords removal, etc.
The algorithms used to perform these operations do not generalize across languages because each language has its own unique features, and are different from one other in terms of their lexical, semantic, and morphological complexities.
While there has been work on data-driven and machine-learned techniques---for example, to stemming~\cite{10.1145/1281485.1281489,6320308,jonker-etal-2020-bag}---for the most part researchers and practitioners have converged on relatively simple and lightweight tokenization pipelines.
For example, in English, the Porter stemmer is widely used, and many systems share stopwords list.

This paper tackles information retrieval in low-resource languages that lack even the most basic language-specific tokenizer.
In this case, the ``default'' and usually the only option would be to simply segment strings into tokens using whitespace.
This obviously is suboptimal, as whitespace-delimited tokens do not capture minor morphological variations that are immaterial from the perspective of search; typically, stemming algorithms would perform this normalization.

How common is this scenario?
One way to characterize the extent of this challenge is to count the number of language-specific tokenizers (called ``analyzers''\footnote{
In the remainder of this paper, we use \textit{analyzer} to refer to  current human-designed tokenization approaches that are language-specific and heuristic-based,
to distinguish from the WordPiece tokenizer discussed throughout this paper. 
}) in the Lucene open-source search library, which underlies search platforms such as Elasticsearch, OpenSearch, and Solr.
As of Lucene 9.3.0, the library provides 42 different language-specific analyzers,\footnote{\url{https://lucene.apache.org/core/9_3_0/analysis/common/index.html}} which cover only a tiny fraction of the commonly cited figure of 7000+ languages that exist on this planet.
It is clear that for most languages, language-specific analyzers don't even exist.

Subword algorithms are actively studied in the context of pretrained language models to alleviate the out-of-vocabulary issue in NLP model training.
Representatives include WordPiece~\citep{Wu2016GooglesNM} and SentencePiece~\citep{kudo-richardson-2018-sentencepiece}.
They are initially applied to English data~\cite{devlin-etal-2019-bert}, then extended to multilingual application scenarios~\cite{devlin-etal-2019-bert, xlmr}. 
For example, the multilingual BERT (mBERT) tokenizer is a WordPiece~\citep{Wu2016GooglesNM} algorithm trained on Wikipedia in 100 languages, comprising 110k vocabulary.

While these tokenizers are actively used as part of  pretrained language models in IR,
they have not been systematically applied independently as an alternative tokenization approach for lexical matching retrieval algorithms. 
However, subword tokenizers have many advantages over expert-designed analyzers and whitespace tokenization.
Compared to analyzers, the tokenizers can be trained automatically, requiring no expert knowledge of the target language.
Additionally, the training is performed on unsupervised data, which could be plentiful even for lower-resource languages.
Compared to whitespace tokenization, we show that subword tokenizers can achieve better retrieval effectiveness.
Moreover, it has been reported that subword tokenizers are less sensitive to misspellings and other irregularities, and can better handle compound words and proper nouns~\citep{zhang-tan-2021-textual}.

In this paper, we propose to improve retrieval for languages that lack language-specific analyzers using the WordPiece tokenizer.
Our results on 11 diverse languages in the Mr.~TyDi test collection~\cite{mrtydi} show that using the WordPiece tokenizer in a lexical matching algorithm consistently outperforms whitespace tokenization, and is even comparable to language-specific Lucene analyzers for some languages.
Additionally, our tokenization approach can be combined with existing Lucene analyzers to further improve retrieval effectiveness. 

\section{Background and Related Work}

Exploring different text representations to improve retrieval in different languages is an active area of research. 
\citet{leveling-et-al} compared different indexing and matching techniques on different levels of abstraction for a document representation in German.
\citet{Jiang2007AnES} explored different tokenization heuristics to improve biomedical information retrieval.

At the same time,
while in recent years IR has widely adopted pretrained models from NLP,
there is little work investigating tokenization in IR tasks using these models. 
To the best of our knowledge, \citet{zhang-tan-2021-textual} is the only work in this line.
They compared different granularities of multilingual textual representations in a traditional IR system (e.g., BM25).
Specifically, they compared the results of tokens produced by the Lucene analyzer in the corresponding language, the SentencePiece~\cite{kudo-richardson-2018-sentencepiece} tokenizer, and directly using characters.
However, the authors focused on the cross-lingual retrieval task and only explored three high-resource languages (i.e., German, French, and Japanese), where SentencePiece tokenization was found to be underwhelming when applied to BM25 alone.

\insertlanguageinfotable
\insertmrtydiresults
\insertmrtydibarplot

\section{Experimental Design}

\textbf{Dataset.}
In this work, we evaluate results on \mrtydi~\cite{mrtydi},
a multilingual retrieval benchmark that extends the TyDi~QA dataset~\cite{tydiqa}.
It provides manually labeled data for monolingual retrieval on 11 typologically diverse languages.
In \mrtydi, queries are questions posed by native speakers of the corresponding languages, and the collection is built from Wikipedia in the same language. 

\smallskip
\noindent
\textbf{Tokenizers.}
We use the BM25 implementation in Anserini~\cite{10.1145/3077136.3080721,10.1145/3239571} with default parameters ($k_1=0.9$, $b=0.4$).
We compare three tokenization mechanisms:\ whitespace, a language-specific Lucene analyzer, and the mBERT tokenizer. 
\autoref{tab:analyzer} provides details on the Lucene analyzers used in work.
Specifically, we use the Porter stemmer and the default stopwords list for English.
We use the mBERT\footnote{\texttt{bert-base-multilingual-uncased} on HuggingFace~\cite{huggingface}}~\cite{devlin-etal-2019-bert} 
tokenizer as the representative of WordPiece tokenization for all languages (including English). 

In \mrtydi, all languages but \Sw have a corresponding custom-built Lucene analyzer, making \Sw the only language that matches the target application scenario in this work (i.e., very low resource). 
However, we still report results using whitespace for all languages, to simulate the situation when the language does not have a language-specific analyzer.
All languages in \mrtydi are included in the training of the mBERT tokenizer. 

Among the three approaches, 
whitespace serves as a ``lower-bound'' baseline,
which is usually less effective but could be applied automatically (although requiring languages that have tokens delimited by whitespace). 
Comparatively, Lucene analyzers are more effective but require expert linguistic knowledge and manual design. 

\smallskip
\noindent
\textbf{Metrics and Fusion.}
We report MRR and recall on the test set of \mrtydi with a cutoff of 100 hits following the original work.
We follow \citet{mrtydi} to perform fusion on retrieval results and refer readers to their paper for details.
However, we do not tune the fusion parameter $\alpha$ in this work.
Instead, we set $\alpha=0.5$ in all conditions. 

\section{Results}

We now examine the effectiveness of the WordPiece tokenizer over languages in different scripts and with diverse typological features.
\autoref{tab:evalres} shows BM25 results with multiple tokenization mechanisms on the test set of \mrtydi:\
(1) whitespace, (2) a language-specific Lucene analyzer, (3) the mBERT tokenizer, and (4) a hybrid of rows~(2,~3).

As the scale of MRR@100 varies across languages,
\autoref{fig:mrtydiMRR} plots the normalized MRR@100 to better demonstrate the relative effectiveness of the tokenization mechanisms. 
For all languages but \Sw (\autoref{fig:mrtydiMRR} left), scores are normalized based on the MRR@100 of the Lucene custom language-specific analyzer.
That is, for each language, the score of the Lucene analyzer is set to 1.0, and the other scores are scaled appropriately. 
For \Sw (\autoref{fig:mrtydiMRR} right), all scores are normalized based on its whitespace score since a language-specific Lucene analyzer is not available.

Comparing results of BM25 using the mBERT tokenizer (row 3) to whitespace (row 1), we observe that the mBERT tokenizer wins on 7 out of 11 languages, including \Sw. 
This is promising, as it indicates that when a Lucene analyzer is not available and whitespace is the only option for tokenization, the WordPiece tokenizer offers a simple approach that potentially yields effective results.
Note that all four cases where the mBERT tokenizer performs worse are on non-Latin-script languages.

Next, comparing the mBERT tokenizer to the language-specific Lucene analyzer (row 2), we observe that the effectiveness gap varies greatly across languages.
For languages such as English (\En) and Russian~(\Ru), BM25 with the mBERT tokenizer obtains similar or even slightly better effectiveness compared to the language-specific Lucene analyzer, and much higher scores compared to whitespace tokenization.
At the other end of the spectrum, for languages such as Telugu (\Te) and Thai (\Th), BM25 with the mBERT tokenizer does not even generate reasonable outputs, yielding scores that are not only substantially worse than using Lucene's custom language-specific analyzer but even whitespace tokenization.

\begin{figure}
    \centering
    \includegraphics[width=0.5\textwidth]{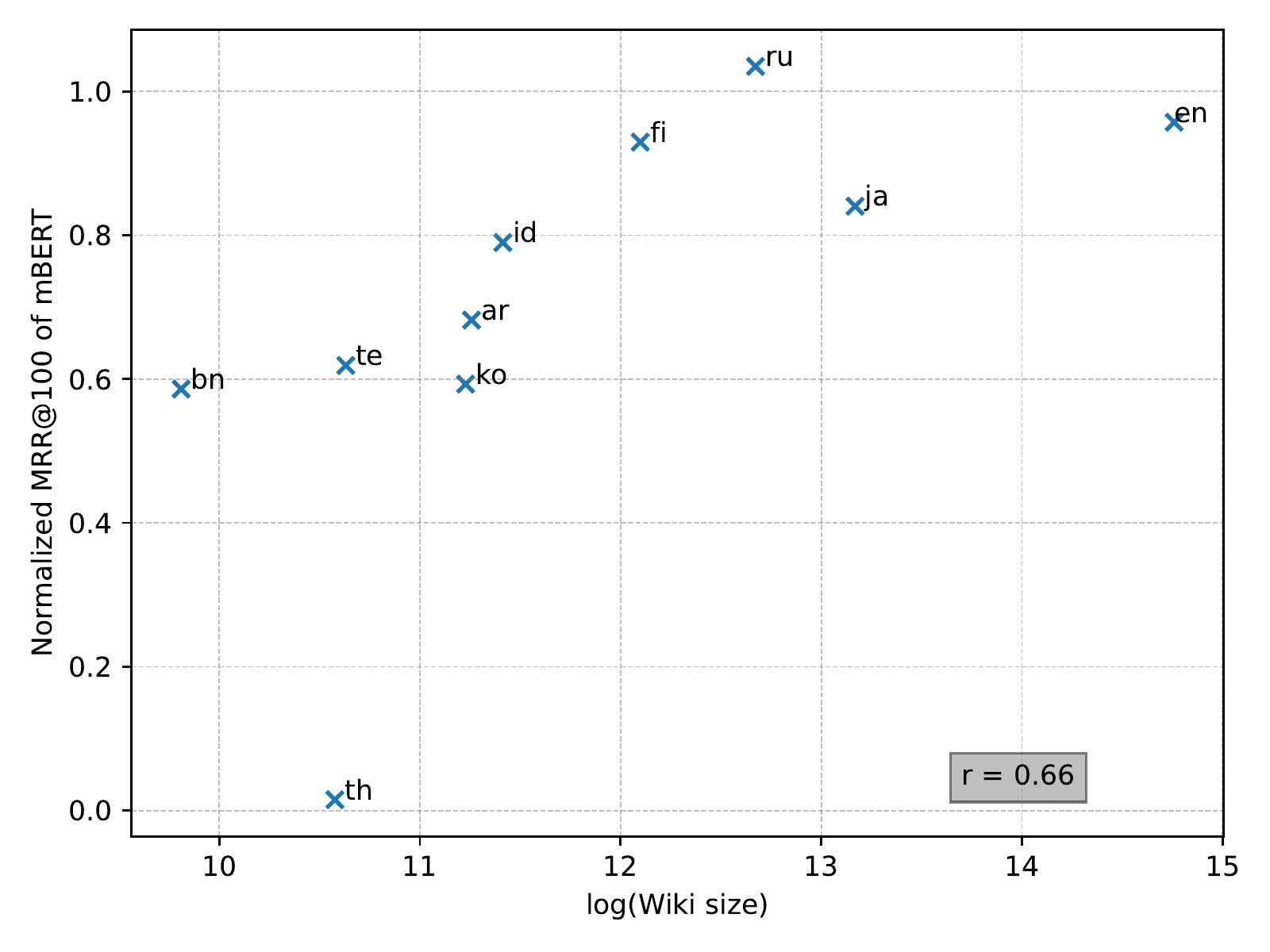}
    \caption{
        A clear correlation between Wikipedia size and the normalized MRR@100 of the mBERT tokenizer.
        All languages but \Sw are included.
        In the figure, $x$-axis shows the Wikipedia size of each language in  $\log$ scale,
        and $y$-axis shows the normalized MRR@100 of mBERT as in \autoref{fig:mrtydiMRR}. 
        The Pearson Correlation coefficient is indicated as $r$.
    }
    \label{fig:delta_col}
\end{figure}

We further explore the reasons behind the largely varied effectiveness gap and found that the relative effectiveness of mBERT to Lucene analyzer is correlated to the size of Wikipedia.
This is shown in \autoref{fig:delta_col},
where the $x$-axis shows the size of Wikipedia\footnote{
We use the ``number of articles'' provided by \url{https://en.wikipedia.org/wiki/Wikipedia:Multilingual_statistics} to indicate the size of Wikipedia in each language.
}
of each language in $\log$ scale,
and the $y$-axis shows the normalized MRR@100 of the mBERT tokenizer as reported in \autoref{fig:mrtydiMRR}.
We use the normalized score since the absolute MRR@100 is affected by multiple factors,
for example, the collection size, the question distribution, etc. 
These factors might differ across languages, but normalization can nicely offset these differences. 
The figure shows a clear correlation between the Wikipedia size and the relative effectiveness of mBERT compared to a custom Lucene analyzer:\ 
the larger the Wikipedia size, the higher the relative effectiveness of mBERT tokenization.

However, as shown by the results of fusing the Lucene analyzer and the mBERT tokenizer, row~(4),
even when mBERT does yield higher scores than the Lucene analyzer by itself,
it generally does not hurt effectiveness,
and even improves the scores in some cases (e.g., \Ru, \Fi, \Ja). 
This finding, however, does not appear to apply to \Te and \Th, where the scores of the mBERT tokenizer are extremely low (see \autoref{tab:evalres}).

\section{Conclusion}

In this paper, we propose to directly apply the WordPiece tokenizer
in lexical matching information retrieval algorithms.
This mechanism is especially useful for low-resource languages that lack language-specific tokenization algorithms handcrafted by language experts.

We evaluate this mechanism on 11 typologically diverse languages.
Results show that splitting text into subwords using the mBERT tokenizer ``out of the box'' provides a promising alternative to whitespace tokenization,
and even beats the custom Lucene analyzer for some languages.
For languages where the mBERT tokenizer achieves good scores, it can be combined with existing Lucene analyzers to provide additional effectiveness gains. 
For low-resource languages that lack effective tokenization algorithms,
we hope this work could be helpful for building robust baselines.

\section*{Acknowledgments}

This research was supported in part by the Natural Sciences and Engineering Research Council (NSERC) of Canada.
Computational resources were provided in part by Compute Ontario and Compute Canada.

\bibliography{custom}

\begin{thebibliography}{16}
\expandafter\ifx\csname natexlab\endcsname\relax\def\natexlab#1{#1}\fi

\bibitem[{Clark et~al.(2020)Clark, Choi, Collins, Garrette, Kwiatkowski,
  Nikolaev, and Palomaki}]{tydiqa}
Jonathan~H. Clark, Eunsol Choi, Michael Collins, Dan Garrette, Tom Kwiatkowski,
  Vitaly Nikolaev, and Jennimaria Palomaki. 2020.
\newblock {T}y{D}i {QA}: A benchmark for information-seeking question answering
  in typologically diverse languages.
\newblock \emph{Transactions of the Association for Computational Linguistics},
  8:454--470.

\bibitem[{Conneau et~al.(2019)Conneau, Khandelwal, Goyal, Chaudhary, Wenzek,
  Guzm{\'{a}}n, Grave, Ott, Zettlemoyer, and Stoyanov}]{xlmr}
Alexis Conneau, Kartikay Khandelwal, Naman Goyal, Vishrav Chaudhary, Guillaume
  Wenzek, Francisco Guzm{\'{a}}n, Edouard Grave, Myle Ott, Luke Zettlemoyer,
  and Veselin Stoyanov. 2019.
\newblock Unsupervised cross-lingual representation learning at scale.
\newblock \emph{arXiv}, abs/1911.02116.

\bibitem[{Devlin et~al.(2019)Devlin, Chang, Lee, and
  Toutanova}]{devlin-etal-2019-bert}
Jacob Devlin, Ming-Wei Chang, Kenton Lee, and Kristina Toutanova. 2019.
\newblock {BERT}: Pre-training of deep bidirectional transformers for language
  understanding.
\newblock In \emph{Proceedings of the 2019 Conference of the North {A}merican
  Chapter of the Association for Computational Linguistics: Human Language
  Technologies, Volume 1 (Long and Short Papers)}, pages 4171--4186,
  Minneapolis, Minnesota.

\bibitem[{Hadni et~al.(2012)Hadni, Lachkar, and Ouatik}]{6320308}
Meryeme Hadni, Abdelmounaime Lachkar, and Sa\"id El~Alaoui Ouatik. 2012.
\newblock A new and efficient stemming technique for {A}rabic text
  categorization.
\newblock In \emph{2012 International Conference on Multimedia Computing and
  Systems}, pages 791--796.

\bibitem[{Jiang and Zhai(2007)}]{Jiang2007AnES}
Jing Jiang and Cheng~Xiang Zhai. 2007.
\newblock An empirical study of tokenization strategies for biomedical
  information retrieval.
\newblock \emph{Information Retrieval}, 10:341--363.

\bibitem[{Jonker et~al.(2020)Jonker, de~Ruijt, and
  de~Gruijl}]{jonker-etal-2020-bag}
Anne Jonker, Corn{\'e} de~Ruijt, and Jornt de~Gruijl. 2020.
\newblock Bag {\&} {T}ag{'}em - a new {D}utch stemmer.
\newblock In \emph{Proceedings of the 12th Language Resources and Evaluation
  Conference}, pages 3868--3876, Marseille, France.

\bibitem[{Kudo and Richardson(2018)}]{kudo-richardson-2018-sentencepiece}
Taku Kudo and John Richardson. 2018.
\newblock {S}entence{P}iece: A simple and language independent subword
  tokenizer and detokenizer for neural text processing.
\newblock In \emph{Proceedings of the 2018 Conference on Empirical Methods in
  Natural Language Processing: System Demonstrations}, pages 66--71, Brussels,
  Belgium.

\bibitem[{Leveling and Hartrumpf(2004)}]{leveling-et-al}
Johannes Leveling and Sven Hartrumpf. 2004.
\newblock University of {H}agen at {CLEF} 2004: Indexing and translating
  concepts for the {GIRT} task.
\newblock In \emph{5th Workshop of the Cross-Language Evaluation Forum}, pages
  271--282.

\bibitem[{Majumder et~al.(2007)Majumder, Mitra, Parui, Kole, Mitra, and
  Datta}]{10.1145/1281485.1281489}
Prasenjit Majumder, Mandar Mitra, Swapan~K. Parui, Gobinda Kole, Pabitra Mitra,
  and Kalyankumar Datta. 2007.
\newblock Yass: {Y}et {A}nother {S}uffix {S}tripper.
\newblock \emph{ACM Trans. Inf. Syst.}, 25(4):18.

\bibitem[{Robertson and Zaragoza(2009)}]{bm25_bibtex}
Stephen Robertson and Hugo Zaragoza. 2009.
\newblock The probabilistic relevance framework: {BM25} and beyond.
\newblock \emph{Found. Trends Inf. Retr.}, 3(4):333–389.

\bibitem[{Wolf et~al.(2020)Wolf, Debut, Sanh, Chaumond, Delangue, Moi, Cistac,
  Rault, Louf, Funtowicz, Davison, Shleifer, von Platen, Ma, Jernite, Plu, Xu,
  Le~Scao, Gugger, Drame, Lhoest, and Rush}]{huggingface}
Thomas Wolf, Lysandre Debut, Victor Sanh, Julien Chaumond, Clement Delangue,
  Anthony Moi, Pierric Cistac, Tim Rault, Remi Louf, Morgan Funtowicz, Joe
  Davison, Sam Shleifer, Patrick von Platen, Clara Ma, Yacine Jernite, Julien
  Plu, Canwen Xu, Teven Le~Scao, Sylvain Gugger, Mariama Drame, Quentin Lhoest,
  and Alexander Rush. 2020.
\newblock Transformers: State-of-the-art natural language processing.
\newblock In \emph{Proceedings of the 2020 Conference on Empirical Methods in
  Natural Language Processing: System Demonstrations}, pages 38--45, Online.

\bibitem[{Wu et~al.(2016)Wu, Schuster, Chen, Le, Norouzi, Macherey, Krikun,
  Cao, Gao, Macherey, Klingner, Shah, Johnson, Liu, Kaiser, Gouws, Kato, Kudo,
  Kazawa, Stevens, Kurian, Patil, Wang, Young, Smith, Riesa, Rudnick, Vinyals,
  Corrado, Hughes, and Dean}]{Wu2016GooglesNM}
Yonghui Wu, Mike Schuster, Z.~Chen, Quoc~V. Le, Mohammad Norouzi, Wolfgang
  Macherey, Maxim Krikun, Yuan Cao, Qin Gao, Klaus Macherey, Jeff Klingner,
  Apurva Shah, Melvin Johnson, Xiaobing Liu, Lukasz Kaiser, Stephan Gouws,
  Yoshikiyo Kato, Taku Kudo, Hideto Kazawa, Keith Stevens, George Kurian,
  Nishant Patil, Wei Wang, Cliff Young, Jason~R. Smith, Jason Riesa, Alex
  Rudnick, Oriol Vinyals, Gregory~S. Corrado, Macduff Hughes, and Jeffrey Dean.
  2016.
\newblock Google's neural machine translation system: Bridging the gap between
  human and machine translation.
\newblock \emph{arXiv}, abs/1609.08144.

\bibitem[{Yang et~al.(2017)Yang, Fang, and Lin}]{10.1145/3077136.3080721}
Peilin Yang, Hui Fang, and Jimmy Lin. 2017.
\newblock Anserini: Enabling the use of {L}ucene for information retrieval
  research.
\newblock In \emph{Proceedings of the 40th International ACM SIGIR Conference
  on Research and Development in Information Retrieval}, SIGIR '17, page
  1253–1256.

\bibitem[{Yang et~al.(2018)Yang, Fang, and Lin}]{10.1145/3239571}
Peilin Yang, Hui Fang, and Jimmy Lin. 2018.
\newblock Anserini: Reproducible ranking baselines using {L}ucene.
\newblock \emph{J. Data and Information Quality}, 10(4).

\bibitem[{Zhang and Tan(2021)}]{zhang-tan-2021-textual}
Hang Zhang and Liling Tan. 2021.
\newblock Textual representations for crosslingual information retrieval.
\newblock In \emph{Proceedings of The 4th Workshop on e-Commerce and NLP},
  pages 116--122, Online.

\bibitem[{Zhang et~al.(2021)Zhang, Ma, Shi, and Lin}]{mrtydi}
Xinyu Zhang, Xueguang Ma, Peng Shi, and Jimmy Lin. 2021.
\newblock Mr. {T}y{D}i: A multi-lingual benchmark for dense retrieval.
\newblock In \emph{Proceedings of the 1st Workshop on Multilingual
  Representation Learning}, pages 127--137, Punta Cana, Dominican Republic.

\end{thebibliography}

\end{document}